\PassOptionsToPackage{table}{xcolor}
\PassOptionsToPackage{mathlines}{lineno}
\documentclass{agujournal2019}
\usepackage{url} 
\usepackage{lineno}
\usepackage[finalnew]{trackchanges} 
\usepackage{soul}
\usepackage{float}
\usepackage{placeins}
\usepackage{amsmath,amssymb,amsfonts}
\usepackage{bm}
\usepackage{pgfplots}
\usepackage{pgfplotstable}
\pgfplotsset{compat=1.16} 
\usepackage{booktabs}
\usepackage{xcolor}
\usepackage{graphicx}
\usepackage{array}
\usepackage{multirow}

\newcommand{\pmcell}[2]{\makebox[4.8em][r]{#1}\,\(\pm\)\,\makebox[4.8em][l]{#2}}
\newif\iflinenumbers
\linenumbersfalse 
\iflinenumbers
\linenumbers
\fi
\let\citep\cite
\let\citet\citeA
\draftfalse

\journalname{JGR: Machine Learning and Computation}

\begin{document}

%
%


\title{LAViG-FLOW: Latent Autoregressive Video Generation for Fluid Flow Simulations}

%
%



\authors{V. De Pellegrini\affil{1}, T. Alkhalifah\affil{1}}

\affiliation{1}{Earth Science and Engineering Program, Physical Science and Engineering Division, King Abdullah University of Science and Technology (KAUST), Thuwal, 23955-6900, Kingdom of Saudi Arabia}

\correspondingauthor{Vittoria De Pellegrini}{vittoria.depellegrini@kaust.edu.sa}


\begin{keypoints}
\item Dedicated autoencoders learn carbon dioxide saturation and pressure build-up fields. Their latents feed a video diffusion transformer model. 
\item This video diffusion transformer model learns their coupled distribution, and shows autoregressive prediction capability. 
\item We demonstrate that this diffusion pipeline is up to two orders of magnitude faster than traditional numerical reservoir solvers.
\end{keypoints}

\begin{abstract}
Modeling and forecasting subsurface multiphase fluid flow fields underpin applications ranging from geological CO\textsubscript{2} sequestration (GCS) operations to geothermal production. This is essential for ensuring both operational performance and long-term safety. While high fidelity multiphase simulators are widely used for this purpose, they become prohibitively expensive once many forward runs are required for inversion purposes and to quantify uncertainty. To tackle this challenge, we propose LAViG-FLOW, a latent autoregressive video generation diffusion framework that explicitly learns the coupled evolution of saturation and pressure fields. Each state variable is compressed by a dedicated 2D autoencoder, and a Video Diffusion Transformer (VDiT) models their coupled distribution across time. We first train the model on a given time horizon to learn their coupled relationship and then fine-tune it autoregressively so it can extrapolate beyond the observed time window. Evaluated on an open-source CO\textsubscript{2} sequestration dataset, LAViG-FLOW generates saturation and pressure fields that stay consistent across time while running two orders of magnitude faster than traditional numerical solvers.
\end{abstract} 

\section*{Plain Language Summary}
Subsurface energy operations such as geological CO\textsubscript{2} sequestration, geothermal production, and hydrogen storage inject fluids into porous rock layers, driving multiphase flow through the connected pore network. These injections change how much pore space is filled (saturation) and how strongly the fluid pushes on the rock (pressure build-up), so engineers must track both fields to avoid fractures or leakage. Traditional reservoir simulators are used to model these coupled effects but are computationally expensive. Because of this, we rely on artificial intelligence and design an alternative to these conventional tools. We build a deep learning workflow, specifically a generative modeling framework, that produces videos of how saturation and pressure fields evolve during CO\textsubscript{2} injection into reservoir formations. We demonstrate that it runs faster than petroleum industry reservoir simulators.  


\section{Introduction}
Modeling and forecasting subsurface multiphase fluid flow fields are needed across a wide range of energy and environmental applications, ranging from geological CO\textsubscript{2} sequestration to geothermal production and hydrogen storage. Accurately monitoring subsurface saturation and pressure evolution is therefore critical for safety and long-term containment. Among these efforts, geological CO\textsubscript{2} sequestration (GCS) draws particular focus because it is a promising strategy to mitigate atmospheric carbon emissions. Supercritical CO\textsubscript{2} is injected into saline aquifers or depleted hydrocarbon formations, migrates under buoyancy, capillarity, and viscous forces, and eventually becomes immobilized via structural, residual, solubility, and mineral trapping mechanisms \cite{bachuCO2StorageGeological2008,leeCO2PlumeMigration2016,saadatpoorNewTrappingMechanism2010a,krevorCapillaryTrappingGeologic2015}.

A conventional numerical simulator, which solves multiphase Darcy flow equations across high-resolution spatial and temporal grids, remains the most robust and widely used tool for predicting subsurface plume migration and pressure evolution \cite{pruessTough2UsersGuide1999,zhaoEfficientSimulationCO22023}. However, its application to realistic scenarios is limited by high computational cost, particularly when multiple realizations are required to do inversion and prediction, as well as quantify the uncertainty involved \cite{nordbottenUncertaintiesPracticalSimulation2012a,ganImpactReservoirParameters2021}.

In the era of artificial intelligence, deep learning models can substitute these costly simulators. While preserving the underlying physics, they can significantly reduce runtime. There are several examples in the literature; among them are transformer-based architectures like TransUNet \cite{tariqDataDrivenMachineLearning2023}, neural operators such as U-FNO and MIONet \cite{wenUFNOAnEnhancedFourier2022,jinMIONetLearningMultipleInput2022}, and enhanced DeepONets \cite{diabUDeepONetUNetEnhanced2024}, which learn complex deterministic mappings from numerical simulator inputs to dynamic field responses.

\begin{figure}[htbp]
    \centering
  \includegraphics[width=\linewidth]{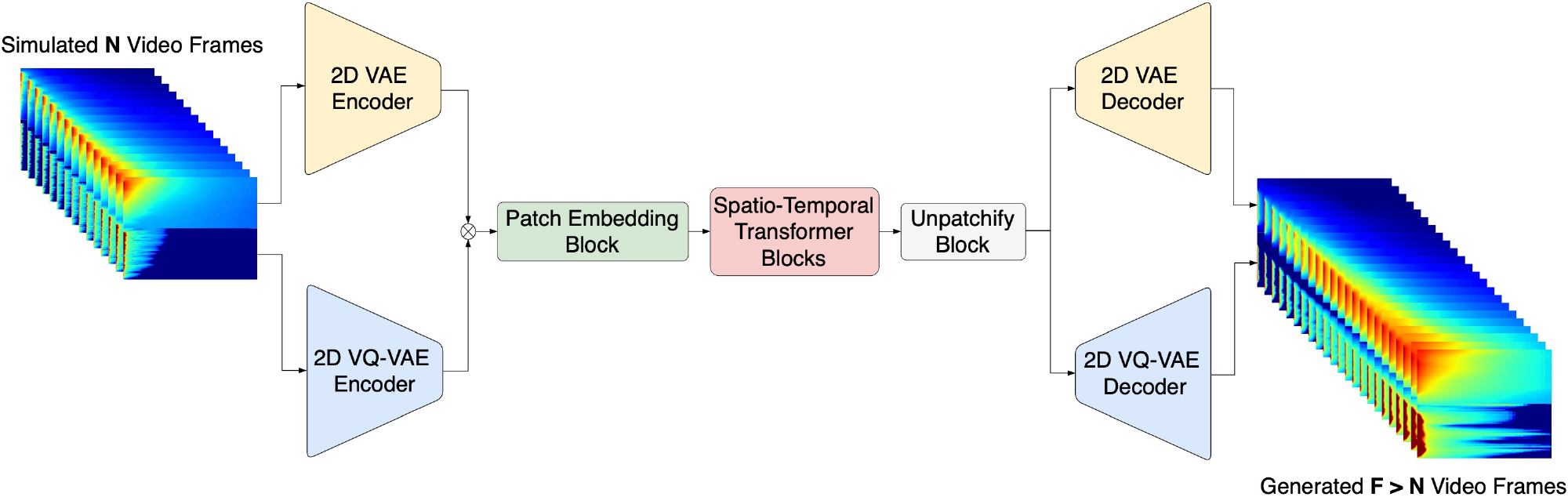}
    \caption{Overview of the LAViG-FLOW pipeline: A reservoir simulator provides $N$ video frames of CO\textsubscript{2} gas saturation and corresponding pressure build-up fields, which are first compressed via 2D VQ-VAE and 2D VAE encoders, concatenated, patch-embedded, and passed through spatio-temporal Transformer blocks before being unpatchified; the VQ-VAE decoder recovers CO\textsubscript{2} gas saturation and the VAE decoder recovers pressure build-up, so the pipeline generates $F{>}N$ video frames at inference.}
    \label{fig:pipeline-overview}
\end{figure}

Here, we explore the use of diffusion models, especially video diffusion models \cite{hoVideoDiffusion2022}, which can address the same task while learning stochastic spatio-temporal data distributions instead of deterministic mappings. \citeA{huangDiffusionCO2Monitoring2024} first proposed a toy example showing how a diffusion model can store the distribution of subsurface elastic-property changes induced by CO\textsubscript{2} injection. Building on this, we introduce LAViG-FLOW, a latent autoregressive video diffusion framework whose DiT backbone builds upon \citeA{maLatte2024} and learns the coupled evolution of CO\textsubscript{2} gas saturation and pressure build-up fields for CO\textsubscript{2} injection scenarios within heterogeneous 2D radially symmetric reservoirs (Figure~\ref{fig:pipeline-overview}).

\subsection{Key Contributions}
\begin{itemize}
    \item We formulate separate latent spaces for physically related state variables, CO\textsubscript{2} gas saturation and pressure build-up, via dedicated 2D Vector Quantized Variational Autoencoder (2D VQ-VAE) and 2D Variational Autoencoder (2D VAE), concatenating their outputs into a shared latent representation for the video diffusion model.
    \item We train a single Video Diffusion Transformer (VDiT) on the shared latent representation formed by concatenating the CO\textsubscript{2} gas saturation and pressure build-up encodings; we demonstrate that this model learns their joint distribution and, after autoregressive fine-tuning, generalizes beyond the training horizon while keeping the outputs physically consistent.
    \item We demonstrate that this diffusion pipeline is up to two orders of magnitude faster than traditional numerical reservoir solvers.
    \item The proposed diffusion pipeline is flexible enough to accept different input sizes and incorporate additional physically related field variables.
\end{itemize}

\section{Related Work}
\subsection{Video Generative Models}
Video generative models have recently achieved remarkable success at generating high quality videos. There are several types of video prediction models that can be distinguished according to their backbone architecture, the dimensional space they operate in, whether they are conditioned and on what kind of conditioning, and other design and training choices. We span this spectrum starting with video predictors that include generative adversarial networks~\cite{goodfellow2014gan} and ending with models that rely on diffusion transformer (DiT) architectures~\cite{peeblesScalableDiffusionModels2022} such as those described as Latte~\cite{maLatte2024} and DiTPainter~\cite{chenDiTPainter2024}. We likewise move from pixel space predictors~\cite{babaeizadeh2021fitvid} toward latent space models \cite{blattmann2023svd,blattmannAlignYourLatents2023}. Pixel space predictors can rely on reconstruction guidance to extend the generation horizon for autoregressive prediction~\cite{ho2022videodiffusion}, and latent space models can rely on autoregressive strategies for long video generation~\cite{zhangPAVDM2024,liuHARP2024}. Conditioning can come from text prompts~\cite{he2022latentvideodiffusion}, 2D scenes~\cite{huang2024movein2d}, or other modalities, and many additional hybrids continue to appear.

\subsection{Video Diffusion Models for Subsurface CO\textsubscript{2} Plumes}
In the field of subsurface fluid flow the only video diffusion example we are aware of is provided by \citeA{huangDiffusionCO2Monitoring2024}, who reuse \citeA{ho2022videodiffusion}'s reconstruction guided pixel space formulation to learn the distribution of elastic-property changes due to CO\textsubscript{2} injection. However, their method operates on fixed low resolution $64\times64$ input frames, works entirely in pixel space, and relies on reconstruction guidance to forecast very short horizons (roughly one additional year) based on early history frames, while evaluation is limited to synthetic, simple scenarios. By contrast, we propose a high resolution latent VDiT model whose backbone architecture is taken from \citeA{maLatte2024} to specifically learn the distribution of CO\textsubscript{2} gas saturation and pressure build-up field variables. The model can accept arbitrary input sizes thanks to its latent space representation and is fine-tuned with autoregressive strategies~\cite{zhangPAVDM2024,liuHARP2024} to generate far-in-the-future frames conditioned on flexible length histories. In addition, our conditioning scheme allows us to choose how many history frames and how many future frames to produce. Finally, the model is evaluated on synthetic data of resolution $96\times200$ that cover a broad range of realistic GCS scenarios \cite{wenUFNOAnEnhancedFourier2022}.

\section{Preliminaries}
\label{sec:preliminaries}

\subsection{Autoregressive Video Prediction Model}
We aim to learn a video prediction model that outputs future frames \(x_{c:F'}=(x_c,\dots,x_{F'-1})\) with \(F'>F\), conditioned on the first \(c\) frames \(x_{<c}=(x_0,\dots,x_{c-1})\) of a video \(x_{0:F-1}=(x_0,\dots,x_{F-1})\), where \(\mathbf{x}_f\in\mathbb{R}^{H\times W\times N_{\text{ch}}}\) is the frame at timestep \(f\) \cite{liuHARP2024, weissenborn2020scaling}. The conditional distribution \(p(x_{c:F'}\mid x_{<c})\) is determined autoregressively in the latent space to accelerate runtime.

\subsection{Variational Autoencoder}
VAE~\cite{kingmaAutoEncodingVariationalBayes2022,higginsBetaVAE2017} pairs an encoder that compresses images into continuous latent representations with a decoder that reconstructs the same image from those latent representations. Given a pressure build-up image $x\in\mathbb{R}^{H\times W\times N_{\text{ch}}}$, the encoder $E$ maps it to $z\in\mathbb{R}^{H'\times W'\times N_z}$ and the decoder $D$ reconstructs it. The VAE is trained by minimizing the following objective function:
\begin{linenomath}
\begin{equation}
\mathcal{L}_{\mathrm{VAE}}(x)
=
\underbrace{\mathbb{E}_{q_\phi(z\mid x)}\left\|\mathcal{D}_\theta(z)-x\right\|_2^2}_{\mathcal{L}_{\mathrm{recon}}}
+\;\beta\;\underbrace{D_{\mathrm{KL}}\!\left(q_\phi(z\mid x)\,\|\,\mathcal{N}(0,\mathbf{I})\right)}_{\mathcal{L}_{\mathrm{KL}}},
\tag{1}
\label{eq:vae_loss}
\end{equation}
\end{linenomath}
where $\mathcal{L}_{\mathrm{recon}}$ is the reconstruction loss that measures the distance between the original input $x$ and its reconstruction $\mathcal{D}_\theta(z)$; the expectation $\mathbb{E}[\cdot]$ inside $\mathcal{L}_{\mathrm{recon}}$ is taken over the encoder distribution $q_\phi(z\mid x)$. $\mathcal{L}_{\mathrm{KL}}$ is the Kullback--Leibler divergence loss weighted by $\beta$, regularizing that encoder distribution toward $\mathcal{N}(0,\mathbf{I})$. 

\subsection{Vector Quantized Variational Autoencoder}
The VQ-VAE~\cite{oord2017neural} pairs an encoder that compresses images into discrete latent representations with a decoder that reconstructs the same image from those latent representations. Given a CO$_2$ gas saturation image $x\in\mathbb{R}^{H\times W\times N_{\text{ch}}}$, the encoder $E$ produces a feature map $z_e(x)\in\mathbb{R}^{H'\times W'\times N_z}$ with spatial size $H'\times W'$. Each of the $H'W'$ spatial locations is a latent vector $z_j(x)\in\mathbb{R}^{N_z}$ for $j=1,\dots,H'W'$. A codebook $\mathcal{C}=\{e_k\}_{k=1}^K$ stores discrete vectors $e_k\in\mathbb{R}^{N_z}$; each latent vector is replaced with its nearest codebook entry, producing the quantized sequence $z_q(x)=(e_{q(x,1)},\dots,e_{q(x,H'W')})$ that the decoder $G$ uses to reconstruct $\hat{x}=G(z_q(x))$. The VQ-VAE is trained by minimizing the following objective:
\begin{linenomath}
\begin{equation}
\mathcal{L}_{\text{VQ-VAE}}(x)
=
\underbrace{\|x-G(z_q(x))\|_2^2}_{\mathcal{L}_{\text{recon}}}
+\underbrace{\|\mathrm{sg}[z_e(x)]-z_q(x)\|_2^2}_{\mathcal{L}_{\text{codebook}}}
+\beta\,\underbrace{\|\mathrm{sg}[z_q(x)]-z_e(x)\|_2^2}_{\mathcal{L}_{\text{commit}}},
\tag{2}
\label{eq:vqvae_loss}
\end{equation}
\end{linenomath}
where $\mathcal{L}_{\text{recon}}$ is again the reconstruction loss, $\mathcal{L}_{\text{codebook}}$ pulls the codebook entries toward the encoder outputs, and $\mathcal{L}_{\text{commit}}$ keeps the encoder stable by encouraging $z_e(x)$ to stay close to the selected code vectors; $\mathrm{sg}[\cdot]$ is the stop-gradient operator that blocks gradients where needed. An optional LPIPS perceptual term $\mathcal{L}_{\text{LPIPS}}$~\cite{zhang2018lpips,johnson2016perceptuallosses} penalizes high-level discrepancies between $x$ and $G(z_q(x))$.

\FloatBarrier
\section{Method}
\label{sec:method}
We present the workflow adopted to train the LAViG-FLOW pipeline (Figure~\ref{fig:pipeline-overview}), which comprises 2 training stages followed by a fine-tuning stage.

\subsection{Stage I: Dual 2D Autoencoder Training}
\begin{figure}[htbp]
    \centering
\includegraphics[width=\linewidth]{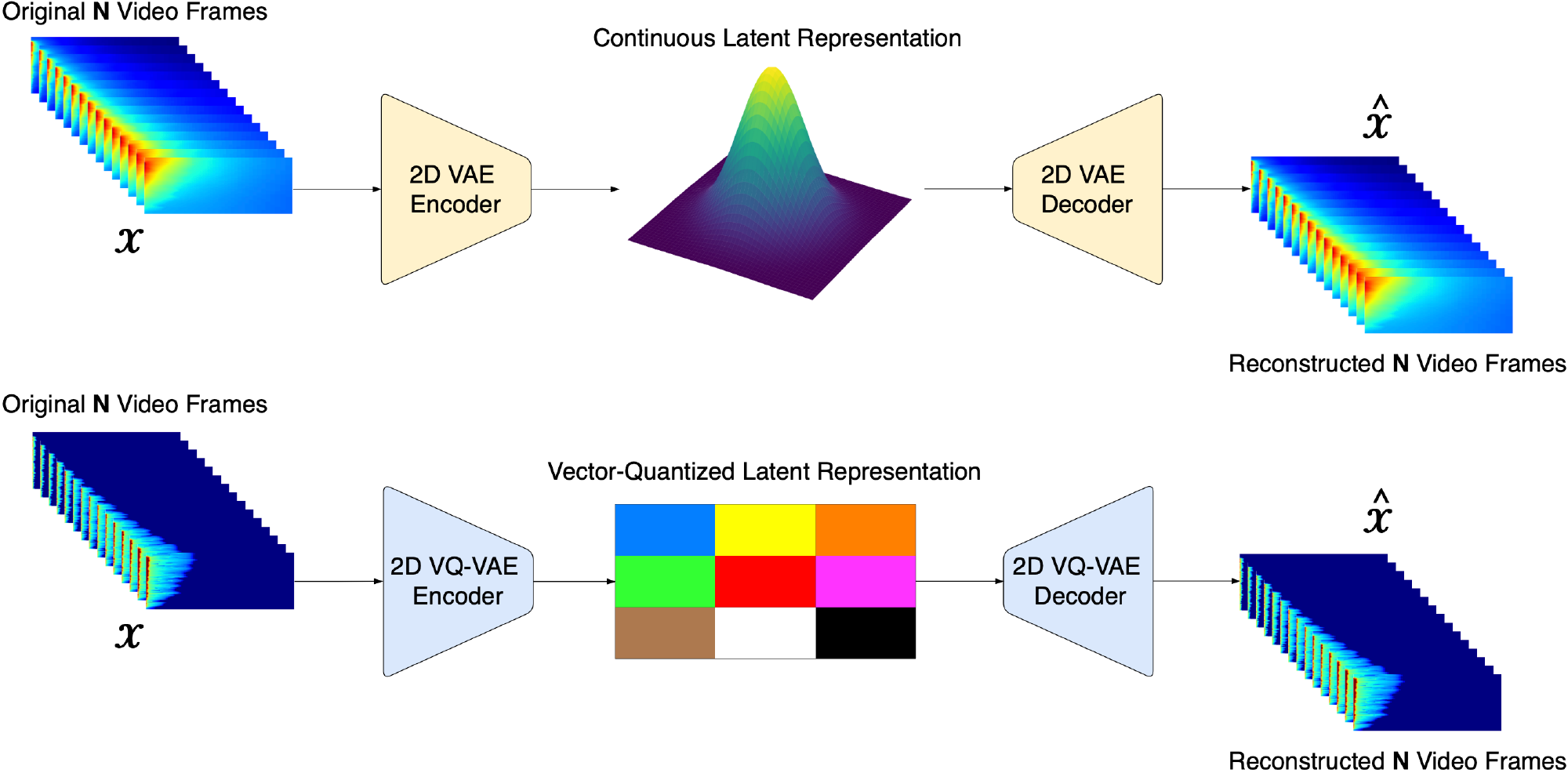}
\caption{Training workflow for \textbf{Stage~I}: the pressure build-up images are compressed into a continuous latent representation, while the CO$_2$ gas saturation images are compressed into a vector-quantized latent representation. Here, $\mathbf{x}$ denotes the input frames and $\hat{\mathbf{x}}$ denotes the reconstructed frames after decoding.}
    \label{fig:vqvae-vae}
\end{figure}
Figure~\ref{fig:vqvae-vae} summarizes the dual path setup used at this stage. We utilize 2D VQ-VAE and VAE models to generate the latent representation for CO$_2$ gas saturation and pressure build-up videos, respectively. Each video clip of length $F$ is reshaped from $\mathbb{R}^{B\times F\times N_{\text{ch}}\times H\times W}$ to $\mathbb{R}^{(B\cdot F)\times N_{\text{ch}}\times H\times W}$; modeling temporal dynamics is left for later diffusion stages. The models are trained using the losses described in Section~\ref{sec:preliminaries}.

\subsection{Stage II: Latent Video Diffusion Transformer Model Pre-Training}
Figure~\ref{fig:stage2and3} summarizes the pre-training workflow of this stage. We pre-train a single latent VDiT model~\cite{maLatte2024} to capture the joint distribution of CO$_2$ gas saturation and pressure build-up clips of length $F$. The input is assembled by concatenating the latents of CO$_2$ gas saturation $z_{\text{gas}}\in\mathbb{R}^{(B\cdot F)\times N_{z_{\text{gas}}}\times H'\times W'}$ and pressure build-up $z_{\text{pressure}}\in\mathbb{R}^{(B\cdot F)\times N_{z_{\text{pressure}}}\times H'\times W'}$ along the channel dimension, yielding $z\in\mathbb{R}^{(B\cdot F)\times(N_{z_{\text{gas}}}+N_{z_{\text{pressure}}})\times H'\times W'}$. We reshape $z$ into $z\in\mathbb{R}^{B\times F\times(N_{z_{\text{gas}}}+N_{z_{\text{pressure}}})\times H'\times W'}$ and feed it to a patch embedding block where every frame is patchified using the uniform frame patch embedding strategy from ViT~\cite{dosovitskiy2021imageisworth}. The resulting tokens are projected to a hidden dimension $D$ and processed by a DiT-style backbone that alternates spatial and temporal self-attention blocks (Variant I)~\cite{maLatte2024}, with AdaLN-Zero conditioning in each block~\cite{peeblesScalableDiffusionModels2022}. After the transformer blocks, tokens are unpatchified into latent frames and decoded back to pixel space. We train the diffusion framework as follows. During the forward diffusion process, we corrupt each clip by sampling $\epsilon\sim\mathcal{N}(0,\mathbf{I})$, defining $\alpha(t)=1-t/T$ for $t\sim\mathcal{U}(0,T)$ with $T=1,000$, and setting $z_t=\alpha(t)\,z_0+(1-\alpha(t))\,\epsilon$. The VDiT then produces two output groups per frame, and we interpret the first group as the velocity field $v_\theta=f_\theta(z_t,t)\in\mathbb{R}^{B\times F\times C\times H'\times W'}$, where $C=N_{z_{\text{gas}}}+N_{z_{\text{pressure}}}$, which we regress toward the clean-minus-noise target:
\begingroup
  \setlength{\abovedisplayskip}{6pt}
  \setlength{\belowdisplayskip}{6pt}
\begin{linenomath}
\begin{equation}
  \mathcal{L}_{\text{RF}} = \mathbb{E}\big[\big\|v_\theta - (z_0 - \epsilon)\big\|_2^2\big],
  \tag{3}
  \label{eq:rectified-flow-loss}
\end{equation}
\end{linenomath}
\endgroup
averaged over spatial and temporal positions; $\mathbb{E}[\cdot]$ denotes the expectation. During the reverse denoising process, the injected noise is iteratively removed by the VDiT predictor, and the reconstruction is implemented via a 30-step sampler that follows rectified-flow theory~\cite{liuRectifiedFlow2023,liu2022flowstraightandfast,kim2025ondevicesora}, instead of the 1,000-step DDPM schedule of \citeA{hoDenoisingDiffusionProbabilistic2020}.

\subsection{Stage III: Autoregressive Latent Video Diffusion Transformer Model Fine-Tuning}
\begin{figure}[htbp]
    \centering
    \includegraphics[width=0.985\linewidth]{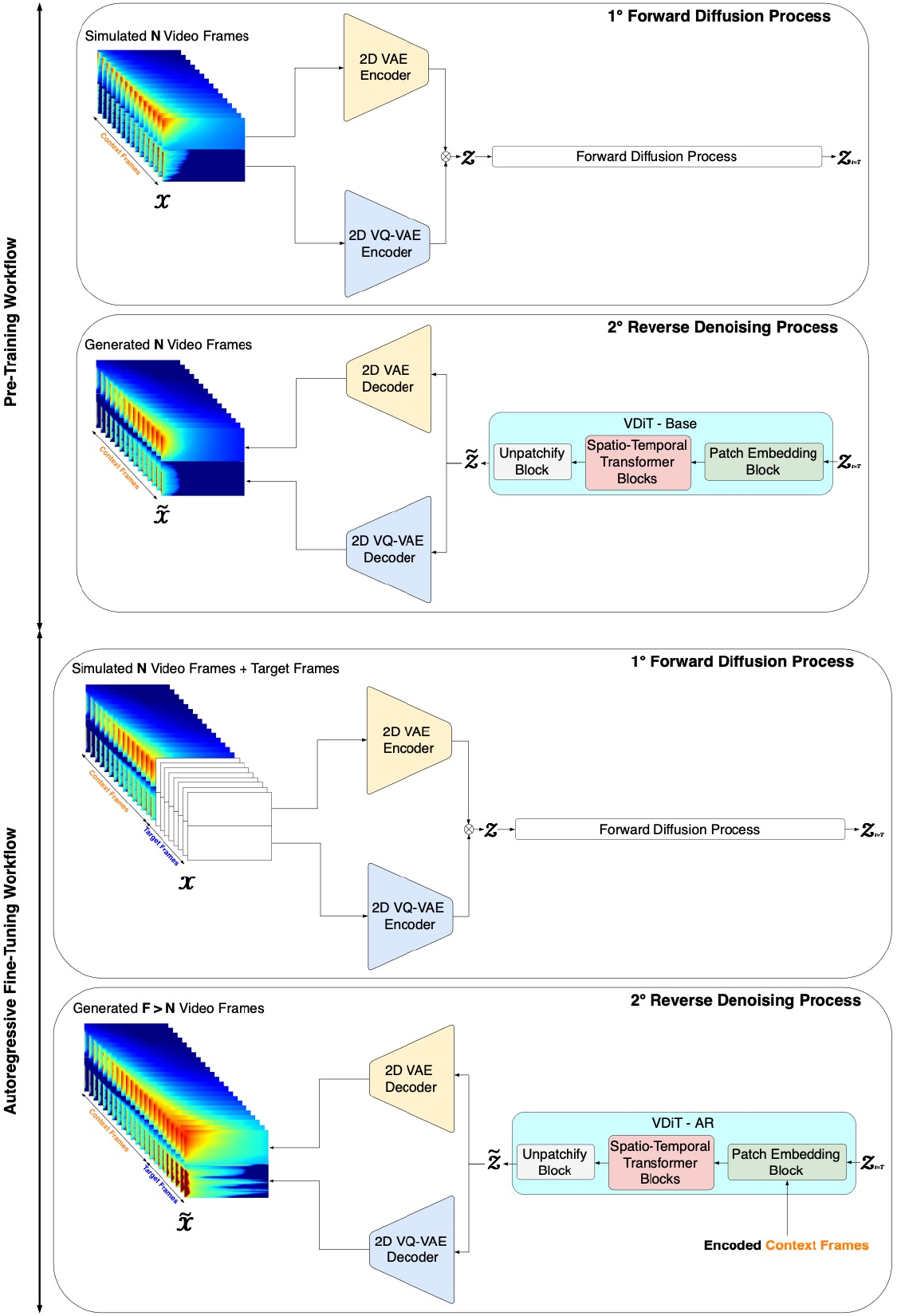}
    \caption{\textbf{Stage~II (pre-training):} Gaussian noise is used to corrupt the joint latent video and VDiT-Base removes it; at inference, VDiT-Base denoises Gaussian noise into an $N$-frame video (same length as input). \textbf{Stage~III (autoregressive fine-tuning):} The same workflow is fine-tuned into VDiT-AR (autoregressive) to predict future frames from $N$ context frames, yielding $F{>}N$ frames; inference is unconditional (start from noise) or conditional (fix encoded context frames). Symbols: $\mathbf{x}$ is the input clip of length $N$ (context frames only, Stage~II) or $N$ + target frames (Stage~III); $\mathbf{z}$ concatenated latents; $\mathbf{z}_{t=T}$ fully noised latents at step $(t{=}T)$; $\tilde{\mathbf{z}}$ denoised latents; $\tilde{\mathbf{x}}$ decoded clip of length $N$ (Stage~II) or $F{>}N$ (Stage~III).}
    \label{fig:stage2and3}
\end{figure}
Figure~\ref{fig:stage2and3} also summarizes the autoregressive fine-tuning workflow. We fine-tune the pre-trained VDiT model to extend the learned distribution of length $F$ videos to longer horizons $F'>F$ for autoregressive prediction. To do this, we define the number of $F_c$ context frames and the number of $F_p$ prediction frames. The context frames represent the observation history, referring to frames within the learned horizon $F$ during Stage II in which the model was trained on, while the prediction frames are those generated beyond the training horizon $F$. We follow the strategy proposed in \citeA{liuHARP2024} and \citeA{zhangPAVDM2024}. Specifically, context latent frames are concatenated with zero initialized placeholders and processed by the VDiT under a binary mask $m\in\{0,1\}^{B\times F}$ that enforces $\mathbf{z}_t \leftarrow m\odot\mathbf{z}_t + (1-m)\odot\mathbf{z}_0$, where $\odot$ denotes the elementwise product, so that context frames stay noise free and only the prediction frames receive noise. This yields a sliding, overlapping procedure: each step treats $F_c$ frames as known and predicts $F_p$, then shifts so the newest $F_c$ frames, combining the recent predictions with the last portion of the preceding context, form the next context window.

\section{Experiments}
We design our experiments to address the following questions:
\begin{itemize}
    \item Can LAViG-FLOW learn the joint distribution of physically coupled state variables (e.g., CO$_2$ gas saturation and pressure build-up)?
    \item Assuming that the answer to the first question is positive, can LAViG-FLOW extrapolate beyond the training horizon via progressive autoregression while maintaining physical consistency?
\end{itemize}
 
\subsection{Implementation}
We train with HuggingFace Accelerate on a GPU cluster (dual Intel\textsuperscript{\textregistered} Xeon\textsuperscript{\textregistered} compute nodes equipped with NVIDIA Tesla V100 GPUs). In Stage I we use 8 V100 GPUs to train the VQ-VAE for 60 epochs with an Adam optimizer at a learning rate of $1\times10^{-5}$ and the VAE for 300 epochs with an Adam optimizer at a learning rate of $1\times10^{-4}$. Both encoders downsample the spatial dimensions by a factor \(f=\frac{H}{h}=\frac{W}{w}=8\) and are frozen for later stages. The VQ-VAE has 3 stride-2 down blocks, 1 mid block, and 3 up blocks (channels $256\!\rightarrow\!384\!\rightarrow\!512\!\rightarrow\!1024$); each block contains 2 residual sub-blocks with convolutions, group norm (32 groups), and SiLU activations, and there are no attention layers. The codebook size is 16{,}384, and the latent channels $z_{\text{gas}}=2$. The VAE mirrors this layout with 3 stride-2 down blocks, one mid block, and 3 up blocks (channels $32\!\rightarrow\!64\!\rightarrow\!128\!\rightarrow\!128$); each block has 1 residual sub-block with convolutions, group norm (16 groups), and SiLU activations, and no attention layers. Latent channels are $z_{\text{pressure}}=24$ (mean/variance are predicted jointly). Stage~II pre-trains the VDiT for 1,000 epochs using an AdamW optimizer with a learning rate of $3\times10^{-5}$, gradient accumulation~$=1$, the 30-step rectified-flow loss described in Section~\ref{sec:method}, a uniform $2\times2$ frame patch embedding, and hidden dimension $D=512$. Stage~III fine-tunes the same backbone for 400 epochs with an autoregressive strategy. Both diffusion stages run on 6 V100 GPUs, and share the VDiT backbone (38,890,192 parameters): 8 transformer layers arranged as alternating spatial and temporal DiT attention blocks (each attention block uses 8 heads with head dimension 64), and timestep embeddings of size 256. 

\subsection{Setup}
We evaluate on the open-source CO\textsubscript{2} sequestration dataset~\cite{wenUFNOAnEnhancedFourier2022}. The dataset is generated with ECLIPSE (E300) simulator on a 2D radially symmetric reservoir of radius 100\,km and discretization $96\times200$ (vertical $\times$ radial), yielding the corresponding $12\times25$ latent grid. Each simulation models 30~years of continuous CO\textsubscript{2} injection at a central vertical well ($r=0.1$\,m) under no-flow vertical boundaries and infinite-acting lateral boundaries. The simulator outputs CO\textsubscript{2} gas saturation and pressure build-up fields over 24 logarithmically spaced timesteps; we use the first 17 frames across all stages (I–III). The full dataset contains 5,500 simulations split into 4,500/500/500 train/validation/test samples. Stage~I trains the autoencoders to produce latents for those 17 frames of CO\textsubscript{2} saturation and pressure build-up. Stage~II pre-trains the VDiT to generate 17-frame videos. In Stage~III we set $F_c=15$ context frames drawn from the same 17 frames and predict $F_p=2$ frames per step; this iterative scheme pushes forecasts beyond the training horizon. In addition, at inference we run 4 progressive steps: after providing $F_c=15$ context frames, stages 1–4 predict 2, 4, 6, and 8 frames, yielding clips of 17, 19, 21, and 23 frames.  We compare these predicted frames against the validation dataset to verify consistency.

\renewcommand{\arraystretch}{1.1}
\setlength{\tabcolsep}{4pt}

For benchmarking, we compare LAViG-FLOW against 5 deterministic surrogates (FNO~\cite{li2020fno}, Conv-FNO~\cite{wen2021ufno}, U-FNO~\cite{wen2021ufno}, Vanilla MIONet~\cite{jinMIONetLearningMultipleInput2022}, and Fourier-MIONet~\cite{jiang2024fouriermionet}). All deterministic baselines are trained using the protocol defined in Table~\ref{tab:baseline_setup}, without additional fine-tuning, and are designed as two separate single-target models (one for CO$_2$ gas saturation and one for pressure build-up $\Delta p$). We compare forecasting performance beyond the 17-frame training horizon by evaluating 2-, 4-, 6-, and 8-frame-ahead predictions.

\begin{table}[H]
\centering
\caption{Benchmark protocol and training settings.}
\label{tab:baseline_setup}
\scriptsize
\resizebox{\textwidth}{!}{%
\begin{tabular}{l c c c c c c c c c c c c}
\toprule
\textbf{Method} & \textbf{Target} & \textbf{Input Space} & \textbf{Frames} & \textbf{Size} & \textbf{Batch Size} & \textbf{Epochs} & \textbf{LR}$^\ddagger$ & \textbf{Step}$^\ddagger$ & \textbf{$\gamma$}$^\ddagger$ & \textbf{Optimizer} & \textbf{Loss Function}$^\dagger$ & \textbf{Precision} \\
\midrule
\multirow{2}{*}{FNO~\cite{li2020fno}} & CO$_2$ & pixel & 17 & 96$\times$200 & 4 & 100 & 1e-3 & 2 & 0.9 & Adam & Rel. $L_2$ + $0.5\nabla_r$ & FP32 \\
 & $\Delta p$ & pixel & 17 & 96$\times$200 & 4 & 140 & 1e-3 & 4 & 0.85 & Adam & Rel. $L_2$ + $0.5\nabla_r$ & FP32 \\
\midrule
\multirow{2}{*}{Conv-FNO~\cite{wen2021ufno}} & CO$_2$ & pixel & 17 & 96$\times$200 & 4 & 100 & 1e-3 & 2 & 0.9 & Adam & Rel. $L_2$ + $0.5\nabla_r$ & FP32 \\
 & $\Delta p$ & pixel & 17 & 96$\times$200 & 4 & 140 & 1e-3 & 4 & 0.85 & Adam & Rel. $L_2$ + $0.5\nabla_r$ & FP32 \\
\midrule
\multirow{2}{*}{U-FNO~\cite{wen2021ufno}} & CO$_2$ & pixel & 17 & 96$\times$200 & 4 & 100 & 1e-3 & 2 & 0.9 & Adam & Rel. $L_2$ + $0.5\nabla_r$ & FP32 \\
 & $\Delta p$ & pixel & 17 & 96$\times$200 & 4 & 140 & 1e-3 & 4 & 0.85 & Adam & Rel. $L_2$ + $0.5\nabla_r$ & FP32 \\
\midrule
\multirow{2}{*}{Vanilla MIONet~\cite{jinMIONetLearningMultipleInput2022}} & CO$_2$ & pixel & 17 & 96$\times$200 & 6 & 150 & 2e-4 & 6 & 0.9 & Adam & Rel. $L_2$ + $0.5\nabla_r$ & FP32 \\
 & $\Delta p$ & pixel & 17 & 96$\times$200 & 6 & 150 & 2e-4 & 6 & 0.9 & Adam & Rel. $L_2$ + $0.5\nabla_r$ & FP32 \\
\midrule
\multirow{2}{*}{Fourier-MIONet~\cite{jiang2024fouriermionet}} & CO$_2$ & pixel & 17 & 96$\times$200 & 4 & 150 & 1e-3 & 3 & 0.9 & Adam & Rel. $L_2$ + $0.5\nabla_r$ & FP32 \\
 & $\Delta p$ & pixel & 17 & 96$\times$200 & 4 & 150 & 1e-3 & 3 & 0.9 & RMSprop & Rel. $L_2$ + $0.5\nabla_r$ & FP32 \\
\midrule
LAViG-FLOW (Ours) -- Pre-Training & CO$_2$ + $\Delta p$ & latent & 17 & 12$\times$25 & 4 & 1000 & 3e-5 & / & / & AdamW & Diffusion (RFlow) & FP16 \\
LAViG-FLOW (Ours) -- Fine-Tuning & CO$_2$ + $\Delta p$ & latent & 17 & 12$\times$25 & 2 & 400 & 3e-5 & / & / & AdamW & Diffusion (RFlow) & FP16 \\
\bottomrule
\end{tabular}%
}
\par\vspace{0.25em}\footnotesize\raggedright$^\ddagger$ LR: learning rate; Step: learning rate step; $\gamma$: learning rate decay factor. $^\dagger$ Rel. $L_2 + 0.5\nabla_r$: two-term loss with a relative $L_2$ output term and a first-derivative regularization term in the radial direction weighted by $\beta=0.5$.
\end{table}

\subsection{Qualitative Analysis}
\begin{figure}[p]
    \centering
  \includegraphics[width=1.0\linewidth]{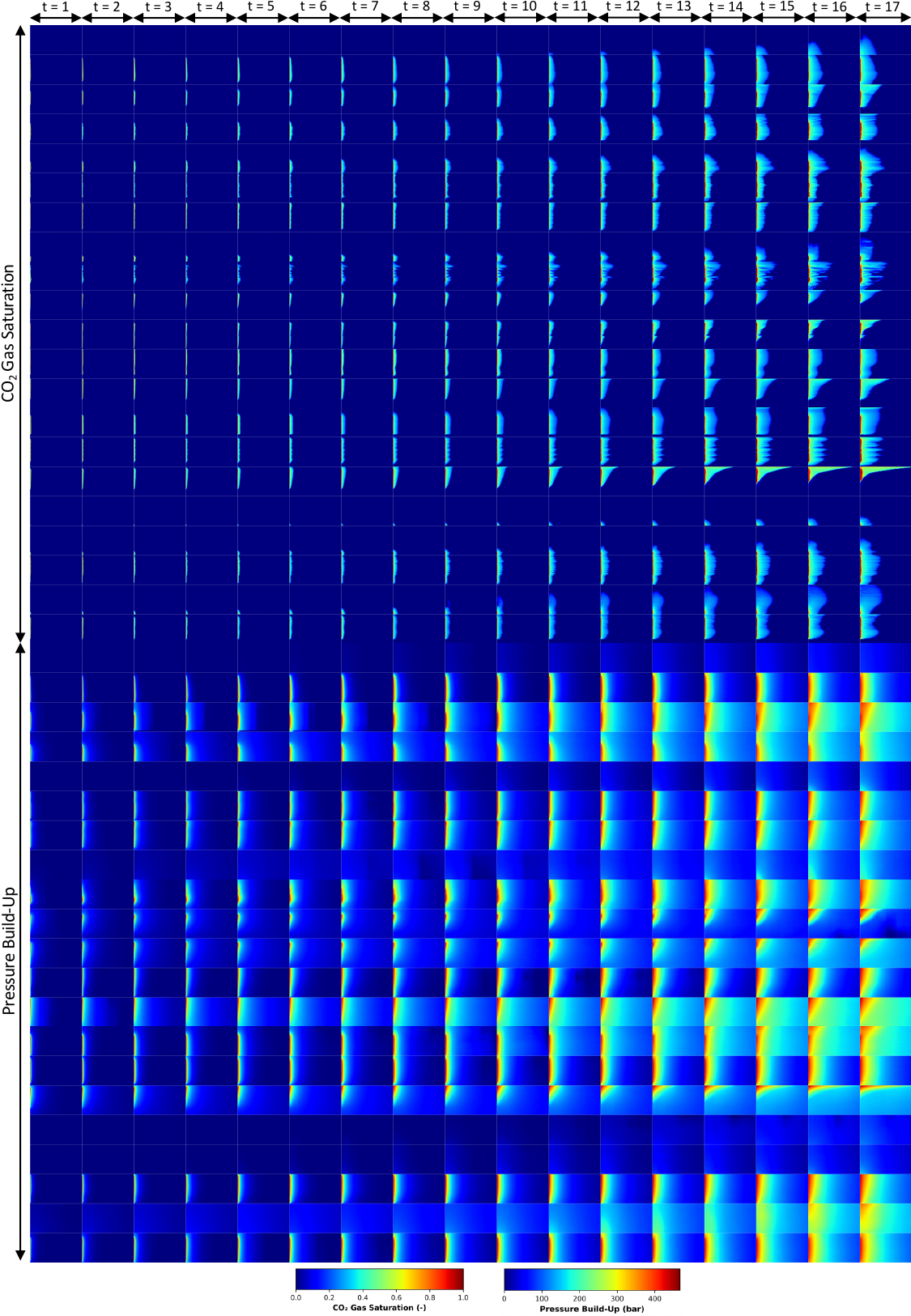}
    \caption{21 unconditionally generated videos from the VDiT model (Stage~II). The top 21 rows show CO\textsubscript{2} gas saturation field across 17 frames; the next 21 rows show the corresponding pressure build-up field, respectively.}
    \label{fig:unconditional}
\end{figure}

\begin{figure}[p]
    \centering
  \vspace*{\fill}
  \includegraphics[width=1.0\linewidth]{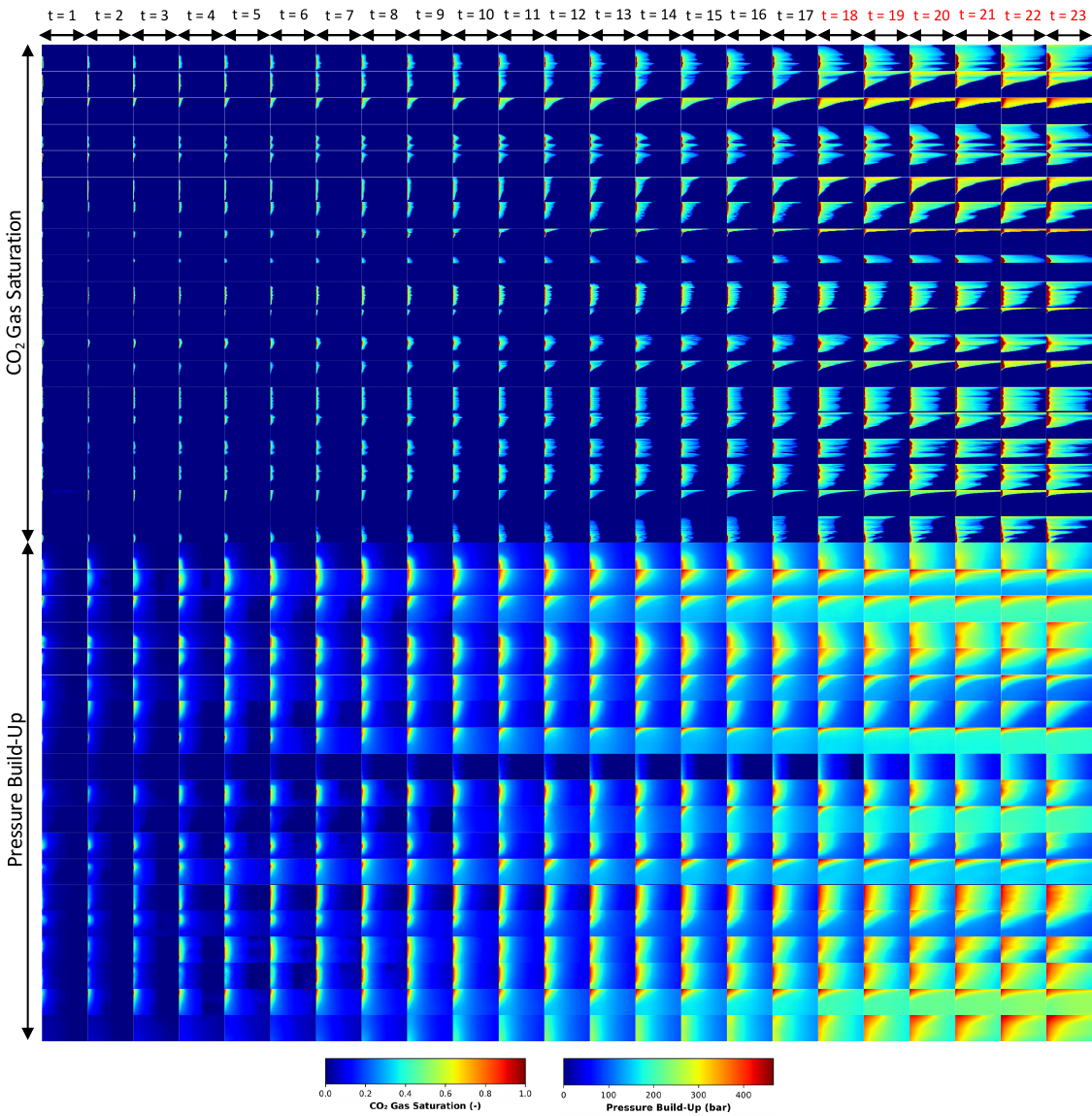}
    \caption{19 unconditionally generated videos using the autoregressive strategy (Stage~III). The top 19 rows show CO\textsubscript{2} gas saturation across 23 frames; the next 19 rows show the corresponding pressure build-up, respectively. Labels for frames 18--23 (beyond the 17-frame training horizon) are shown in red.}
    \label{fig:autoregressive}
  \vspace*{\fill}
\end{figure}

\begin{figure}[p]
    \centering
  \includegraphics[width=1.0\linewidth]{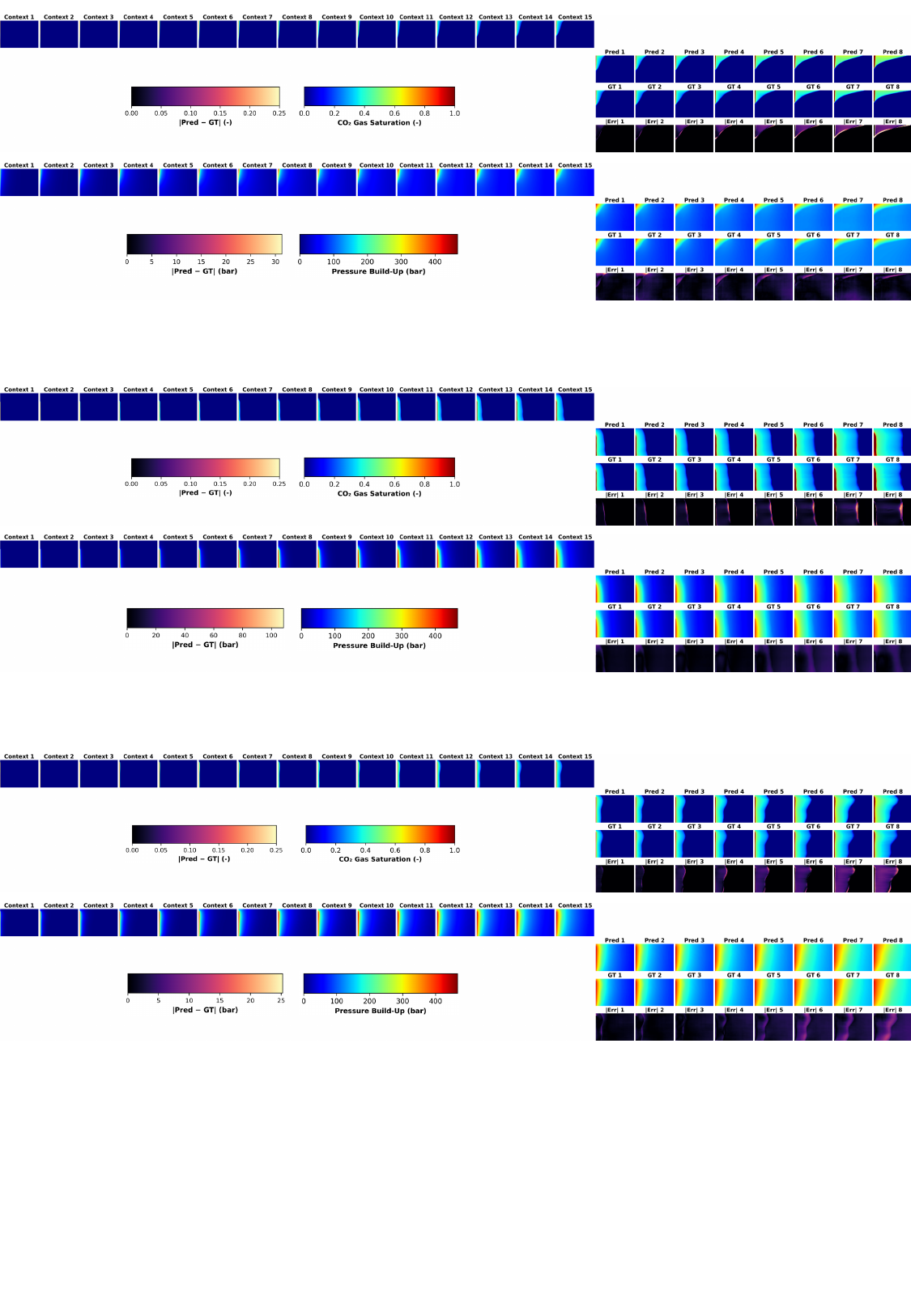}
    \caption{3 conditionally sampled videos using $F_c=15$ context frames from the validation set (Stage~III inference). Each case includes the total $F_p=8$ predicted frames, the corresponding ground truth, and the absolute error between them.}
    \label{fig:conditional}
\end{figure}

Figure~\ref{fig:unconditional} shows 21 unconditionally generated videos by the VDiT (Stage~II), each depicting CO\textsubscript{2} saturation and pressure build-up over 17 frames. Figure~\ref{fig:autoregressive} (Stage~III) presents 19 unconditionally generated videos using the autoregressive strategy. Figure~\ref{fig:conditional} (Stage~III inference) plots 3 sampled videos conditioned on $F_c=15$ frames; stages 1–4 predict 2, 4, 6, and 8 frames (23 total), and we compare the predictions with the corresponding ground truths. By visually inspecting the generated videos we can make the following observations. For Stage~II, the unconditionally generated videos exhibit smooth radial plume expansion and both CO\textsubscript{2} saturation and pressure build-up fields evolve together, confirming the hidden correspondence learned during joint training; all scenarios match the training setup, with injections starting on the left side where the well is located. Moving to Stage~III, the autoregressive strategy extends this behavior beyond frame 17 without breaking the link between the two fields, and the red-labeled frames remain coherent with their preceding histories. Finally, the Stage~III conditional results confirm that, with 15 observed frames, the predicted frames closely match the ground-truth validation clips, with no significant discrepancies between predictions and references. 

\subsection{Quantitative Analysis}
Table~\ref{tab:autoregression_metrics} reports reconstruction metrics (MSE, mean squared error; MAE, mean absolute error; RMSE, root mean squared error), and Table~\ref{tab:quality_metrics} reports video quality metrics (SSIM, structural similarity index; PSNR, peak signal-to-noise ratio; LPIPS, learned perceptual image patch similarity; FVD, Fr\'echet video distance) for both CO$_2$ and $\Delta p$ across methods and prediction stages. Since Stage~4 corresponds to the longest forecasting horizon (8 future frames), it is the most challenging and most informative stage for comparison. At this horizon, LAViG-FLOW achieves the best metrics for both targets, with lower reconstruction errors and better video quality scores than all deterministic baselines.
Table~\ref{tab:generation_time} summarizes per-sample inference time across CPU/GPU settings and reports Stage~4 CPU 1 core speed-up relative to ECLIPSE, expressed as a multiplicative factor ($\times$, i.e., times). As expected, deterministic surrogates are faster per stage because they require a single forward pass, whereas LAViG-FLOW is slower due to latent encoding/decoding and 30 diffusion denoising steps. Nevertheless, for joint CO$_2$ + $\Delta p$ video generation, LAViG-FLOW remains faster than full ECLIPSE simulation (approximately 575 s for 23 frames, linearly scaled from about 600 s for 24 frames~\cite{wenUFNOAnEnhancedFourier2022}), with a Stage~4 speed-up of approximately $2.69\times$.
Table~\ref{tab:params_train_time} compares model size, required number of GPUs, and training time. Compared with deterministic models, LAViG-FLOW has the highest total training cost and parameter count because it jointly models two coupled variables (CO$_2$ and $\Delta p$) within a single generative framework.

\clearpage

\FloatBarrier
\begin{table}[H]
\centering
\caption{Reconstruction metrics (mean $\pm$ std) averaged over 500 validation samples across prediction stages (videos for LAViG-FLOW; multi-step predictions for deterministic baselines). Best values across methods at Stage~4 (8-frame-ahead) are \textbf{bolded} and highlighted in green.}
\label{tab:autoregression_metrics}
\tiny
\renewcommand{\arraystretch}{1.10}
\setlength{\tabcolsep}{1.5pt}
\resizebox{\linewidth}{!}{%
\begin{tabular}{l c c c c c c c c}
\toprule
\textbf{Method} & \textbf{Stage} & \textbf{Pred.\ frames} & \multicolumn{3}{c}{\textbf{CO$_2$}} & \multicolumn{3}{c}{\textbf{$\bm{\Delta p}$}} \\
\cmidrule(lr){4-6}\cmidrule(lr){7-9}
 &  &  & \textbf{MSE$\downarrow$} & \textbf{MAE$\downarrow$} & \textbf{RMSE$\downarrow$} & \textbf{MSE$\downarrow$} & \textbf{MAE$\downarrow$} & \textbf{RMSE$\downarrow$} \\
\midrule
\multirow{4}{*}{FNO~\cite{li2020fno}} & 1 & 2 & \pmcell{0.028235}{0.023532} & \pmcell{0.065293}{0.031083} & \pmcell{0.155820}{0.062892} & \pmcell{0.340050}{1.017998} & \pmcell{0.235232}{0.259111} & \pmcell{0.405676}{0.418900} \\
 & 2 & 4 & \pmcell{0.083050}{0.056503} & \pmcell{0.130493}{0.053147} & \pmcell{0.272223}{0.094575} & \pmcell{0.281963}{0.850365} & \pmcell{0.230147}{0.219707} & \pmcell{0.384129}{0.366617} \\
 & 3 & 6 & \pmcell{0.262180}{0.081565} & \pmcell{0.342040}{0.063629} & \pmcell{0.505427}{0.081996} & \pmcell{0.247097}{0.480488} & \pmcell{0.252446}{0.152255} & \pmcell{0.428846}{0.251373} \\
 & 4 & 8 & \pmcell{1.026718}{0.137743} & \pmcell{0.828150}{0.053807} & \pmcell{1.011217}{0.064480} & \pmcell{0.296122}{0.453077} & \pmcell{0.284845}{0.153578} & \pmcell{0.487170}{0.242461} \\
\midrule
\multirow{4}{*}{Conv-FNO~\cite{wen2021ufno}} & 1 & 2 & \pmcell{0.012453}{0.011414} & \pmcell{0.034762}{0.018890} & \pmcell{0.102098}{0.045046} & \pmcell{0.627087}{2.587404} & \pmcell{0.269587}{0.375816} & \pmcell{0.470653}{0.636846} \\
 & 2 & 4 & \pmcell{0.031676}{0.022712} & \pmcell{0.075841}{0.032311} & \pmcell{0.168254}{0.058024} & \pmcell{0.990212}{4.106694} & \pmcell{0.359051}{0.512739} & \pmcell{0.581241}{0.807695} \\
 & 3 & 6 & \pmcell{0.152307}{0.084244} & \pmcell{0.236787}{0.080287} & \pmcell{0.376063}{0.104325} & \pmcell{1.078587}{4.533813} & \pmcell{0.393897}{0.517405} & \pmcell{0.628587}{0.826720} \\
 & 4 & 8 & \pmcell{1.069525}{0.260292} & \pmcell{0.739445}{0.117062} & \pmcell{1.026199}{0.128219} & \pmcell{0.701716}{1.642201} & \pmcell{0.376378}{0.336565} & \pmcell{0.640027}{0.540445} \\
\midrule
\multirow{4}{*}{U-FNO~\cite{wen2021ufno}} & 1 & 2 & \pmcell{0.005252}{0.008437} & \pmcell{0.011862}{0.011455} & \pmcell{0.057605}{0.043974} & \pmcell{4.399755}{20.109449} & \pmcell{0.424728}{0.786196} & \pmcell{0.968358}{1.860655} \\
 & 2 & 4 & \pmcell{0.008654}{0.011352} & \pmcell{0.017660}{0.014442} & \pmcell{0.079429}{0.048422} & \pmcell{20.954350}{76.433723} & \pmcell{1.009382}{1.841511} & \pmcell{2.203479}{4.012360} \\
 & 3 & 6 & \pmcell{0.016319}{0.013737} & \pmcell{0.030144}{0.016265} & \pmcell{0.118870}{0.046782} & \pmcell{20.955994}{78.181358} & \pmcell{1.177635}{2.136713} & \pmcell{2.193055}{4.018271} \\
 & 4 & 8 & \pmcell{0.032323}{0.016928} & \pmcell{0.051988}{0.018245} & \pmcell{0.174453}{0.043467} & \pmcell{10.262449}{36.270218} & \pmcell{0.955800}{1.580758} & \pmcell{1.628118}{2.758927} \\
\midrule
\multirow{4}{*}{Vanilla MIONet~\cite{jinMIONetLearningMultipleInput2022}} & 1 & 2 & \pmcell{0.041837}{0.025874} & \pmcell{0.075509}{0.024408} & \pmcell{0.195747}{0.059331} & \pmcell{0.085942}{0.088293} & \pmcell{0.145526}{0.088879} & \pmcell{0.265291}{0.124752} \\
 & 2 & 4 & \pmcell{0.043415}{0.023725} & \pmcell{0.078148}{0.021910} & \pmcell{0.201367}{0.053540} & \pmcell{0.080429}{0.077612} & \pmcell{0.149159}{0.079193} & \pmcell{0.259671}{0.114017} \\
 & 3 & 6 & \pmcell{0.049918}{0.022808} & \pmcell{0.086302}{0.021101} & \pmcell{0.218084}{0.048553} & \pmcell{0.080764}{0.069935} & \pmcell{0.159327}{0.072050} & \pmcell{0.264537}{0.103848} \\
 & 4 & 8 & \pmcell{0.059488}{0.024060} & \pmcell{0.096575}{0.023386} & \pmcell{0.239185}{0.047735} & \pmcell{0.085387}{0.064363} & \pmcell{0.171642}{0.067681} & \pmcell{0.276139}{0.095574} \\
\midrule
\multirow{4}{*}{Fourier-MIONet~\cite{jiang2024fouriermionet}} & 1 & 2 & \pmcell{0.069619}{0.047577} & \pmcell{0.095220}{0.040011} & \pmcell{0.249370}{0.086220} & \pmcell{0.217742}{0.304856} & \pmcell{0.218802}{0.164207} & \pmcell{0.395987}{0.246854} \\
 & 2 & 4 & \pmcell{0.064689}{0.040950} & \pmcell{0.094528}{0.034519} & \pmcell{0.242676}{0.076141} & \pmcell{0.205182}{0.288837} & \pmcell{0.218385}{0.154894} & \pmcell{0.385247}{0.238257} \\
 & 3 & 6 & \pmcell{0.066905}{0.035249} & \pmcell{0.100496}{0.029891} & \pmcell{0.250524}{0.064366} & \pmcell{0.198796}{0.274180} & \pmcell{0.225880}{0.144674} & \pmcell{0.383606}{0.227250} \\
 & 4 & 8 & \pmcell{0.073459}{0.031081} & \pmcell{0.109040}{0.027982} & \pmcell{0.265301}{0.055442} & \pmcell{0.197198}{0.259931} & \pmcell{0.235915}{0.135080} & \pmcell{0.388561}{0.214983} \\
\midrule
\multirow{4}{*}{\textbf{LAViG-FLOW (Ours)}} & 1 & 2 & \pmcell{0.001535}{0.001834} & \pmcell{0.007593}{0.005329} & \pmcell{0.034466}{0.018632} & \pmcell{0.000940}{0.001517} & \pmcell{0.010392}{0.008744} & \pmcell{0.024830}{0.017980} \\
 & 2 & 4 & \pmcell{0.002754}{0.003046} & \pmcell{0.010643}{0.007168} & \pmcell{0.046782}{0.023771} & \pmcell{0.001686}{0.002483} & \pmcell{0.014250}{0.012600} & \pmcell{0.033583}{0.023634} \\
 & 3 & 6 & \pmcell{0.004457}{0.004584} & \pmcell{0.014168}{0.009333} & \pmcell{0.060034}{0.029208} & \pmcell{0.002864}{0.004082} & \pmcell{0.019146}{0.016618} & \pmcell{0.044054}{0.030380} \\
 & \textbf{4} & \textbf{8} & \cellcolor{green!10}\textbf{\pmcell{0.006750}{0.006745}} & \cellcolor{green!10}\textbf{\pmcell{0.018425}{0.012152}} & \multicolumn{1}{c@{\hspace{0.6em}}}{\cellcolor{green!10}\textbf{\pmcell{0.074236}{0.035207}}} & \cellcolor{green!10}\textbf{\pmcell{0.004624}{0.006594}} & \cellcolor{green!10}\textbf{\pmcell{0.025081}{0.021908}} & \cellcolor{green!10}\textbf{\pmcell{0.056482}{0.037867}} \\
\bottomrule
\end{tabular}%
}
\end{table}
\vspace{-0.45em}

\begin{table}[H]
\centering
\caption{Video quality metrics (mean $\pm$ std) averaged over 500 validation samples across prediction stages (videos for LAViG-FLOW; multi-step predictions for deterministic baselines). Best values across methods at Stage~4 (8-frame-ahead) are \textbf{bolded} and highlighted in green.}
\label{tab:quality_metrics}
\tiny
\renewcommand{\arraystretch}{1.10}
\setlength{\tabcolsep}{1.4pt}
\resizebox{\linewidth}{!}{%
\begin{tabular}{l c c c c c c c c c c}
\toprule
\textbf{Method} & \textbf{Stage} & \textbf{Pred.\ frames} & \multicolumn{4}{c}{\textbf{CO$_2$}} & \multicolumn{4}{c}{\textbf{$\bm{\Delta p}$}} \\
\cmidrule(lr){4-7}\cmidrule(lr){8-11}
 &  &  & \textbf{SSIM$\uparrow$} & \textbf{PSNR$\uparrow$} & \textbf{LPIPS$\downarrow$} & \textbf{FVD$\downarrow$} & \textbf{SSIM$\uparrow$} & \textbf{PSNR$\uparrow$} & \textbf{LPIPS$\downarrow$} & \textbf{FVD$\downarrow$} \\
\midrule
\multirow{4}{*}{FNO~\cite{li2020fno}} & 1 & 2 & \pmcell{0.773832}{0.051126} & \pmcell{22.826574}{3.345830} & \pmcell{0.256369}{0.053473} & \pmcell{99.947678}{4.001024} & \pmcell{0.552025}{0.178861} & \pmcell{16.165094}{5.533671} & \pmcell{0.444984}{0.115580} & \pmcell{70.349844}{1.559171} \\
 & 2 & 4 & \pmcell{0.443185}{0.110976} & \pmcell{17.858456}{3.092790} & \pmcell{0.503536}{0.051910} & \pmcell{197.298955}{3.403555} & \pmcell{0.416921}{0.140536} & \pmcell{16.116398}{4.707772} & \pmcell{0.539531}{0.065013} & \pmcell{150.185955}{3.669017} \\
 & 3 & 6 & \pmcell{0.278028}{0.058489} & \pmcell{12.069807}{1.488092} & \pmcell{0.592967}{0.032613} & \pmcell{504.747314}{9.932213} & \pmcell{0.261510}{0.112711} & \pmcell{14.222285}{3.478778} & \pmcell{0.606436}{0.049259} & \pmcell{190.248988}{5.109212} \\
 & 4 & 8 & \pmcell{0.295702}{0.042762} & \pmcell{5.940421}{0.532215} & \pmcell{0.600800}{0.022439} & \pmcell{2831.749245}{27.303380} & \pmcell{0.154612}{0.096489} & \pmcell{12.993907}{3.372623} & \pmcell{0.624260}{0.067545} & \pmcell{235.063002}{8.415816} \\
\midrule
\multirow{4}{*}{Conv-FNO~\cite{wen2021ufno}} & 1 & 2 & \pmcell{0.861297}{0.045894} & \pmcell{26.638943}{3.718287} & \pmcell{0.181467}{0.045818} & \pmcell{50.183044}{3.886443} & \pmcell{0.559825}{0.198295} & \pmcell{15.787152}{6.349249} & \pmcell{0.439443}{0.118777} & \pmcell{99.298627}{2.094157} \\
 & 2 & 4 & \pmcell{0.575481}{0.091983} & \pmcell{21.985954}{2.873607} & \pmcell{0.428831}{0.051959} & \pmcell{159.583953}{2.715780} & \pmcell{0.430074}{0.187679} & \pmcell{14.167857}{6.549708} & \pmcell{0.522981}{0.082710} & \pmcell{245.301813}{8.659164} \\
 & 3 & 6 & \pmcell{0.133888}{0.062593} & \pmcell{14.854534}{2.450751} & \pmcell{0.646484}{0.031357} & \pmcell{661.470138}{12.454714} & \pmcell{0.303290}{0.162196} & \pmcell{12.987500}{5.882402} & \pmcell{0.588045}{0.057599} & \pmcell{462.246232}{20.714136} \\
 & 4 & 8 & \pmcell{0.048397}{0.023730} & \pmcell{5.866781}{1.123602} & \pmcell{0.700308}{0.018062} & \pmcell{1123.875241}{25.770738} & \pmcell{0.205232}{0.125110} & \pmcell{11.633625}{4.928460} & \pmcell{0.615710}{0.066818} & \pmcell{602.632843}{22.740495} \\
\midrule
\multirow{4}{*}{U-FNO~\cite{wen2021ufno}} & 1 & 2 & \pmcell{0.964167}{0.027376} & \pmcell{32.913975}{5.903586} & \pmcell{0.061369}{0.030840} & \pmcell{4.931860}{0.464662} & \pmcell{0.671001}{0.164068} & \pmcell{14.032584}{10.656052} & \pmcell{0.314337}{0.109457} & \pmcell{42.488588}{2.569797} \\
 & 2 & 4 & \pmcell{0.916985}{0.044963} & \pmcell{29.422443}{4.859748} & \pmcell{0.136897}{0.059086} & \pmcell{8.415718}{0.474104} & \pmcell{0.298889}{0.144039} & \pmcell{8.236431}{11.688020} & \pmcell{0.505098}{0.075677} & \pmcell{121.882464}{2.818433} \\
 & 3 & 6 & \pmcell{0.799841}{0.068767} & \pmcell{25.140907}{3.262272} & \pmcell{0.254428}{0.063404} & \pmcell{27.287024}{1.570684} & \pmcell{0.327330}{0.160141} & \pmcell{7.847807}{11.185040} & \pmcell{0.498845}{0.082538} & \pmcell{132.744811}{3.384970} \\
 & 4 & 8 & \pmcell{0.682391}{0.094381} & \pmcell{21.446861}{2.119992} & \pmcell{0.348376}{0.067002} & \pmcell{93.094007}{4.943946} & \pmcell{0.325688}{0.161590} & \pmcell{8.701134}{9.680935} & \pmcell{0.493148}{0.092350} & \pmcell{104.527275}{4.643709} \\
\midrule
\multirow{4}{*}{Vanilla MIONet~\cite{jinMIONetLearningMultipleInput2022}} & 1 & 2 & \pmcell{0.820207}{0.049345} & \pmcell{20.585503}{2.655907} & \pmcell{0.194629}{0.032469} & \pmcell{37.005657}{2.431830} & \pmcell{0.659323}{0.169412} & \pmcell{18.426481}{3.909312} & \pmcell{0.295145}{0.101395} & \pmcell{16.580081}{0.284018} \\
 & 2 & 4 & \pmcell{0.811615}{0.048642} & \pmcell{20.243429}{2.300687} & \pmcell{0.203247}{0.034395} & \pmcell{44.127900}{3.420919} & \pmcell{0.629848}{0.156314} & \pmcell{18.488979}{3.597789} & \pmcell{0.310104}{0.098126} & \pmcell{22.617633}{0.892933} \\
 & 3 & 6 & \pmcell{0.791973}{0.048318} & \pmcell{19.459883}{1.920114} & \pmcell{0.220357}{0.037306} & \pmcell{46.045553}{3.632303} & \pmcell{0.577255}{0.136520} & \pmcell{18.177259}{3.216845} & \pmcell{0.331828}{0.092033} & \pmcell{20.982710}{0.536784} \\
 & 4 & 8 & \pmcell{0.769465}{0.051678} & \pmcell{18.618465}{1.737667} & \pmcell{0.240729}{0.041885} & \pmcell{53.442299}{6.623531} & \pmcell{0.526951}{0.119374} & \pmcell{17.683079}{2.895715} & \pmcell{0.356291}{0.085136} & \pmcell{22.440114}{0.765823} \\
\midrule
\multirow{4}{*}{Fourier-MIONet~\cite{jiang2024fouriermionet}} & 1 & 2 & \pmcell{0.807529}{0.041349} & \pmcell{18.624001}{3.131764} & \pmcell{0.198632}{0.031603} & \pmcell{42.443469}{3.020608} & \pmcell{0.646314}{0.158955} & \pmcell{15.569416}{5.110434} & \pmcell{0.357194}{0.099901} & \pmcell{27.298214}{0.808029} \\
 & 2 & 4 & \pmcell{0.796358}{0.041317} & \pmcell{18.751211}{2.770005} & \pmcell{0.207328}{0.032893} & \pmcell{48.639793}{3.329072} & \pmcell{0.613554}{0.150977} & \pmcell{15.731833}{4.910524} & \pmcell{0.369162}{0.100442} & \pmcell{36.946384}{1.582320} \\
 & 3 & 6 & \pmcell{0.774330}{0.034547} & \pmcell{18.326185}{2.223486} & \pmcell{0.226382}{0.031798} & \pmcell{56.440803}{4.679581} & \pmcell{0.564021}{0.131086} & \pmcell{15.601878}{4.556684} & \pmcell{0.387614}{0.097164} & \pmcell{38.641839}{0.783345} \\
 & 4 & 8 & \pmcell{0.750322}{0.031279} & \pmcell{17.736753}{1.835234} & \pmcell{0.248580}{0.031548} & \pmcell{67.916791}{6.220007} & \pmcell{0.516408}{0.111601} & \pmcell{15.301472}{4.169700} & \pmcell{0.409366}{0.091863} & \pmcell{34.692014}{1.157032} \\
\midrule
\multirow{4}{*}{\textbf{LAViG-FLOW (Ours)}} & 1 & 2 & \pmcell{0.975930}{0.021077} & \pmcell{35.365922}{4.718814} & \pmcell{0.022635}{0.017042} & \pmcell{2.662142}{0.750915} & \pmcell{0.982947}{0.030614} & \pmcell{39.128845}{6.098365} & \pmcell{0.023597}{0.024557} & \pmcell{0.899870}{0.213468} \\
 & 2 & 4 & \pmcell{0.959621}{0.033599} & \pmcell{31.419064}{4.514990} & \pmcell{0.039182}{0.026934} & \pmcell{6.225630}{1.075178} & \pmcell{0.971415}{0.042239} & \pmcell{35.467939}{6.203025} & \pmcell{0.031298}{0.032306} & \pmcell{1.313884}{0.249193} \\
 & 3 & 6 & \pmcell{0.938537}{0.050023} & \pmcell{28.506751}{4.378132} & \pmcell{0.057042}{0.037415} & \pmcell{7.180662}{1.105508} & \pmcell{0.953147}{0.057751} & \pmcell{32.281809}{6.385566} & \pmcell{0.045493}{0.044619} & \pmcell{2.332724}{0.272063} \\
 & \textbf{4} & \textbf{8} & \cellcolor{green!10}\textbf{\pmcell{0.909094}{0.074901}} & \cellcolor{green!10}\textbf{\pmcell{26.115494}{4.230441}} & \cellcolor{green!10}\textbf{\pmcell{0.077476}{0.048501}} & \multicolumn{1}{c@{\hspace{0.6em}}}{\cellcolor{green!10}\textbf{\pmcell{11.851343}{1.781195}}} & \cellcolor{green!10}\textbf{\pmcell{0.927854}{0.076077}} & \cellcolor{green!10}\textbf{\pmcell{29.217966}{6.142274}} & \cellcolor{green!10}\textbf{\pmcell{0.063788}{0.054780}} & \cellcolor{green!10}\textbf{\pmcell{3.159129}{0.340041}} \\
\bottomrule
\end{tabular}%
}
\end{table}
\vspace{-0.45em}

\begin{table}[H]
\centering
\caption{Per-sample inference time across CPU/GPU settings and prediction stages (video generation for LAViG-FLOW; multi-step predictions for deterministic baselines). Speed-up (Stage~4, CPU 1 core) is relative to ECLIPSE. Lowest times and highest speed-up are \textbf{bolded} and highlighted in green.}
\label{tab:generation_time}
\tiny
\renewcommand{\arraystretch}{1.10}
\setlength{\tabcolsep}{1.2pt}
\resizebox{\linewidth}{!}{%
\begin{tabular}{l c c c c c c c c c c c c c c}
\toprule
\textbf{Method} & \textbf{Target} & \multicolumn{4}{c}{\textbf{CPU 1 core (s)}} & \multicolumn{1}{c}{\textbf{Speed-up (×)}} & \multicolumn{4}{c}{\textbf{CPU 4 cores (s)}} & \multicolumn{4}{c}{\textbf{V100 (s)}} \\
\cmidrule(lr){3-6}\cmidrule(lr){7-7}\cmidrule(lr){8-11}\cmidrule(lr){12-15}
 &  & \shortstack{\textbf{Stage 1}\\\textbf{17f}} & \shortstack{\textbf{Stage 2}\\\textbf{19f}} & \shortstack{\textbf{Stage 3}\\\textbf{21f}} & \shortstack{\textbf{Stage 4}\\\textbf{23f}} & \shortstack{\textbf{Stage 4}\\\textbf{23f}} & \shortstack{\textbf{Stage 1}\\\textbf{17f}} & \shortstack{\textbf{Stage 2}\\\textbf{19f}} & \shortstack{\textbf{Stage 3}\\\textbf{21f}} & \shortstack{\textbf{Stage 4}\\\textbf{23f}} & \shortstack{\textbf{Stage 1}\\\textbf{17f}} & \shortstack{\textbf{Stage 2}\\\textbf{19f}} & \shortstack{\textbf{Stage 3}\\\textbf{21f}} & \shortstack{\textbf{Stage 4}\\\textbf{23f}} \\
\midrule
\multirow{3}{*}{\textbf{FNO}~\textbf{\cite{li2020fno}}} & CO$_2$ & 9.6535 & 10.5332 & 12.8855 & 14.2201 & -- & 3.9182 & 4.1824 & 4.6836 & 4.9131 & 0.0279 & 0.0285 & 0.0316 & 0.0304 \\
 & $\Delta p$ & 10.7849 & 11.7614 & 13.8084 & 14.7930 & -- & 3.9364 & 4.1949 & 4.6177 & 4.9152 & 0.0279 & 0.0285 & 0.0316 & 0.0304 \\
 & $\bm{\mathrm{CO_2} + \Delta p}$ & \cellcolor{green!10}\textbf{20.4383} & \cellcolor{green!10}\textbf{22.2946} & \cellcolor{green!10}\textbf{26.6939} & \cellcolor{green!10}\textbf{29.0131} & \cellcolor{green!10}\textbf{19.82x} & \cellcolor{green!10}\textbf{7.8546} & \cellcolor{green!10}\textbf{8.3773} & \cellcolor{green!10}\textbf{9.3013} & \cellcolor{green!10}\textbf{9.8283} & \cellcolor{green!10}\textbf{0.0559} & \cellcolor{green!10}\textbf{0.0569} & \cellcolor{green!10}\textbf{0.0632} & \cellcolor{green!10}\textbf{0.0608} \\
\midrule
\multirow{3}{*}{Conv-FNO~\cite{wen2021ufno}} & CO$_2$ & 32.1540 & 34.4913 & 36.7687 & 34.5021 & -- & 7.9174 & 8.5387 & 9.0870 & 9.0861 & 0.0459 & 0.0479 & 0.0525 & 0.0528 \\
 & $\Delta p$ & 16.5242 & 17.2807 & 18.8484 & 16.7636 & -- & 5.3020 & 5.5096 & 5.8603 & 5.9230 & 0.0459 & 0.0478 & 0.0524 & 0.0527 \\
 & CO$_2$ + $\Delta p$ & 48.6783 & 51.7721 & 55.6171 & 51.2657 & 11.22x & 13.2194 & 14.0484 & 14.9473 & 15.0091 & 0.0919 & 0.0957 & 0.1049 & 0.1055 \\
\midrule
\multirow{3}{*}{U-FNO~\cite{wen2021ufno}} & CO$_2$ & 83.7210 & 91.2743 & 99.2795 & 106.3672 & -- & 19.2260 & 20.9601 & 22.8560 & 24.3865 & 0.0807 & 0.0850 & 0.0925 & 0.0957 \\
 & $\Delta p$ & 22.7589 & 24.8195 & 26.9824 & 28.8118 & -- & 6.6175 & 7.1325 & 7.8607 & 8.2890 & 0.0805 & 0.0853 & 0.0923 & 0.0955 \\
 & CO$_2$ + $\Delta p$ & 106.4799 & 116.0937 & 126.2618 & 135.1790 & 4.25x & 25.8436 & 28.0927 & 30.7167 & 32.6755 & 0.1612 & 0.1703 & 0.1848 & 0.1912 \\
\midrule
\multirow{3}{*}{Vanilla MIONet~\cite{jinMIONetLearningMultipleInput2022}} & CO$_2$ & 43.8434 & 48.9961 & 54.1337 & 59.3081 & -- & 9.5841 & 10.6524 & 11.8177 & 12.9530 & 0.1190 & 0.1401 & 0.1523 & 0.1665 \\
 & $\Delta p$ & 34.8448 & 38.9276 & 43.0345 & 47.1578 & -- & 7.5476 & 8.4125 & 9.3541 & 10.2397 & 0.1215 & 0.1431 & 0.1494 & 0.1635 \\
 & CO$_2$ + $\Delta p$ & 78.6883 & 87.9237 & 97.1682 & 106.4659 & 5.40x & 17.1317 & 19.0649 & 21.1719 & 23.1927 & 0.2404 & 0.2832 & 0.3017 & 0.3300 \\
\midrule
\multirow{3}{*}{Fourier-MIONet~\cite{jiang2024fouriermionet}} & CO$_2$ & 34.6805 & 38.9178 & 42.7991 & 46.8078 & -- & 7.9608 & 8.8577 & 9.7682 & 10.7195 & 0.0351 & 0.0392 & 0.0432 & 0.0470 \\
 & $\Delta p$ & 36.8648 & 41.3462 & 45.3466 & 49.7050 & -- & 8.6670 & 9.7404 & 10.7271 & 11.7785 & 0.0351 & 0.0391 & 0.0432 & 0.0469 \\
 & CO$_2$ + $\Delta p$ & 71.5452 & 80.2639 & 88.1457 & 96.5128 & 5.96x & 16.6278 & 18.5981 & 20.4953 & 22.4981 & 0.0702 & 0.0783 & 0.0863 & 0.0939 \\
\midrule
\multirow{1}{*}{LAViG-FLOW (Ours)} & CO$_2$ + $\Delta p$ & 133.9957 & 160.6537 & 187.3160 & 213.4959 & 2.69x & 32.6118 & 39.2213 & 45.6319 & 52.4643 & 0.8690 & 1.1622 & 1.4498 & 1.7381 \\
\bottomrule
\end{tabular}%
}
\end{table}
\vspace{-0.45em}

\begin{table}[H]
\centering
\caption{Model size, required number of GPUs, and total training time across methods. The lowest total training times are \textbf{bolded} and highlighted in green.}
\label{tab:params_train_time}
\tiny
\renewcommand{\arraystretch}{0.95}
\setlength{\tabcolsep}{1.8pt}
\resizebox{\textwidth}{!}{%
\begin{tabular}{l c c c >{\centering\arraybackslash}p{4.2cm}}
\toprule
\textbf{Method} & \textbf{Target} & \textbf{N. Parameters} & \textbf{N. GPUs} & \multicolumn{1}{c}{\textbf{Total Training Time (d/h/m/s)}} \\
\midrule
\multirow{2}{*}{FNO~\cite{li2020fno}} & CO$_2$ & 31,117,325 & 2 $\times$ Tesla V100-SXM2-32GB & 7h 27m 7s \\
 & $\Delta p$ & 31,117,325 & 2 $\times$ Tesla V100-SXM2-32GB & 10h 53m 8s \\
\midrule
\multirow{2}{*}{Conv-FNO~\cite{wen2021ufno}} & CO$_2$ & 31,222,409 & 2 $\times$ Tesla V100-SXM2-32GB & 13h 57m 24s \\
 & $\Delta p$ & 31,222,409 & 2 $\times$ Tesla V100-SXM2-32GB & 19h 44m 11s \\
\midrule
\multirow{2}{*}{U-FNO~\cite{wen2021ufno}} & CO$_2$ & 32,164,709 & 2 $\times$ Tesla V100-SXM2-32GB & 1d 4h 15m 3s \\
 & $\Delta p$ & 32,164,709 & 2 $\times$ Tesla V100-SXM2-32GB & 1d 15h 34m 35s \\
\midrule
\multirow{2}{*}{\textbf{Vanilla MIONet~\cite{jinMIONetLearningMultipleInput2022}}} & \textbf{CO$_2$} & 10,424,769 & 1 $\times$ Tesla V100-SXM2-32GB & \cellcolor{green!10}\textbf{5h 5m 8s} \\
 & \textbf{$\bm{\Delta p}$} & 10,424,769 & 1 $\times$ Tesla V100-SXM2-32GB & \cellcolor{green!10}\textbf{3h 21m 4s} \\
\midrule
\multirow{2}{*}{Fourier-MIONet~\cite{jiang2024fouriermionet}} & CO$_2$ & 3,685,325 & 1 $\times$ Tesla V100-SXM2-32GB & 12h 27m 8s \\
 & $\Delta p$ & 3,685,325 & 1 $\times$ Tesla V100-SXM2-32GB & 13h 40m 15s \\
\midrule
\multirow{2}{*}{LAViG-FLOW (Ours)} & \multirow{2}{*}{CO$_2$ + $\Delta p$} & 38,890,192 & \multirow{2}{*}{6 $\times$ Tesla V100-SXM2-32GB} & Pre-Training: 3d 18h 52m 6s \\
 &  & 38,890,192 &  & Fine-Tuning: 1d 15h 8m 32s \\
\bottomrule
\end{tabular}%
}
\end{table}

\clearpage
\section{Discussion \& Future Work}
Finally, we share the limitations of the approach and possible extensions. First, each temporal attention layer in the VDiT uses fixed absolute positional embeddings: it knows frame 2 follows frame 1, but it is not aware that frame 2 might correspond to “year 5.” Switching to rotary positional embeddings would let us express time offsets between frames and request a specific time frame (e.g., “generate year 10”).

Second, we evaluate on a sparse set of timesteps: frames are far apart in time, so the model misses the evolution of CO\textsubscript{2} saturation and pressure. Denser time sampling would reveal the coupled dynamics more clearly and help the model learn them. Conditioning is also limited to historical context frames, leaving additional physics unmodeled. A planned extension involves conditioning on physical controls (e.g., injection rates) to guide generation and to test whether the model can forecast for specified injection scenarios.

Lastly, the pre-training stage (Stage II) is costly, taking about 3 days to complete. Mixed precision and batch size strongly affect memory use and computational cost: higher precision increases memory use and smaller batches increase runtime. We trained the VQ-VAE and VAE in full precision (float32) with batch size 4. On top of these latent representations we pre-train and fine-tune the VDiT in lower precision (float16), with patch size 2, batch size 4 for pre-training, and batch size 2 for fine-tuning. To reduce memory and speed up runtime, we can move the autoencoders to lower precision (e.g., bfloat16 or float16), increase batch sizes, and increase patch size to cut the number of tokens. Finding the right configuration requires some tuning: for instance, increasing the patch size can reduce spatial resolution, and increasing the batch size can destabilize optimization or hurt generalization.

\section{Conclusion}
We presented LAViG-FLOW, a latent autoregressive video generation diffusion model that learns the coupled evolution of CO\textsubscript{2} gas saturation and pressure build-up in multiphase flow scenarios. It learns their joint distribution and generates high-resolution videos showing their combined evolution; with autoregressive fine-tuning, it extends the learned horizon from length $F$ to $F'$, demonstrating forecasting ability.

Quantitative analysis shows that at Stage~4 (8-frame-ahead), which is the hardest forecasting setting, LAViG-FLOW consistently provides the strongest reconstruction and video quality scores for both CO\textsubscript{2} and $\Delta p$, while deterministic surrogates degrade more clearly at this horizon. This gain in prediction quality comes with higher model complexity, including larger parameter count, longer total training time, and higher generation time per sample. Even with this cost, for joint CO$_2$ + $\Delta p$ video generation, LAViG-FLOW remains faster than full ECLIPSE simulation (approximately 575 seconds per 23-frame sample on an Intel\textsuperscript{\textregistered} Xeon\textsuperscript{\textregistered} Processor E5-2670 CPU), corresponding to a Stage~4 (CPU 1 core) speed-up of about $2.69\times$.



%
%

%
%
%


\section*{Open Research}
The CO\textsubscript{2} geological sequestration simulation dataset used for training and evaluation is available as described in \citeA{wenUFNOAnEnhancedFourier2022}. The LAViG-FLOW training and inference code, together with the trained checkpoints, are available at \url{https://github.com/DeepWave-KAUST/LAViG-FLOW-pub}. The full pipeline is implemented in Python, and all figures shown in this manuscript were generated using Matplotlib 3.9.4~\cite{hunterMatplotlib2DGraphics2007}.

\acknowledgments
This publication is based on work supported by the King Abdullah University of Science and Technology (KAUST). The authors gratefully acknowledge the support of the DeepWave sponsors and the computational resources provided by KAUST's high-performance computing facilities. Special thanks are extended to the developers of the GCS open-source dataset used in this study.

\section*{Conflicts of Interest}
The authors declare there are no conflicts of interest for this manuscript.

\bibliography{Bibliography}

\end{document}